\documentclass[10pt]{article}

\usepackage[american]{babel}

\usepackage{natbib} 
\usepackage{mathtools} 
\usepackage{graphicx,subcaption,booktabs,hyperref}
\usepackage{bbm}
\usepackage{booktabs} 
\usepackage{tikz} 
\usetikzlibrary{shapes, arrows.meta, positioning, fit, calc, intersections, decorations.pathreplacing, tikzmark, backgrounds}

\newcommand{\bs}[1]{\boldsymbol{#1}}
\tikzset{
    nodeStyle/.style={align=center, minimum height=2em, minimum width=3cm}
}

\newcommand{\DataTable}[5]{

    \pgfmathsetmacro{\width}{3*#3}
    \pgfmathsetmacro{\height}{5*#3}
    \pgfmathsetmacro{\shiftx}{#1 - \width/2}
    \pgfmathsetmacro{\shifty}{#2 - \height/2}
    
    \begin{scope}[shift={(\shiftx,\shifty)}, scale=#3]
    \draw   (0,0) rectangle (3,5);
    \draw    (0,4) -- (3, 4);
    \draw  [dash pattern={on 1.5pt off 1.5pt}]  (0,3-0.05) -- (3, 3-0.05);
    \draw  [dash pattern={on 1.5pt off 1.5pt}]  (0,2-0.05) -- (3, 2-0.05);
    \draw  [dash pattern={on 1.5pt off 1.5pt}]  (0,1-0.05) -- (3, 1-0.05);
    \ifx\\#4\\%
    \else
        \fill[#4, opacity=0.2] (0,3) rectangle (3,4);
    \fi
    \ifx\\#5\\%
    \else
        \fill[#5, opacity=0.2] (0,2) rectangle (3,3);
    \fi
    \draw (1.5,4.5) node  {$\mathcal{D}$};
    \draw (1.5,3.5) node  {$x_{1}$};
    \draw (1.5,2.5) node  {${\textstyle x_{2}}$};
    \draw (1.5,1.7) node  {$\vdots $};
    \draw (1.5,0.5) node  {$x_{n}$};
    \end{scope}
}
\usepackage{amsmath}
\usepackage{amssymb}
\usepackage{mathtools}
\usepackage{amsthm}
\usepackage[ruled]{algorithm2e}
\usepackage{authblk}

\usepackage[capitalize,noabbrev]{cleveref}

\theoremstyle{plain}
\newtheorem{theorem}{Theorem}[section]

\theoremstyle{definition}
\newtheorem{definition}[theorem]{Definition}

\theoremstyle{remark}

\DeclareMathOperator{\argmin}{argmin}

\title{On the Usage of Gaussian Process for Efficient Data Valuation}
\author[a]{Clément Bénesse,}
\author[b,c]{Patrick Mesana,}
\author[d]{Athénaïs Gautier,}
\author[b]{Sébastien Gambs}

\affil[a]{Opsci.ai\\
Paris, France}
\affil[b]{Université du Québec à Montréal\\
Montréal, Québec, Canada}
\affil[c]{HEC Montréal,\\
Montréal, Québec, Canada}
\affil[d]{COSMO - Stochastic Mine Planning Laboratory, Department of Mining and Materials Engineering\\
McGill University, Montreal, Quebec, Canada}
\date{}

\begin{document}

\maketitle

\vspace*{0.2in}

\begin{abstract}
In machine learning, knowing the impact of a given datum on model training is a fundamental task referred to as Data Valuation.
Building on previous works from the literature, we have designed a novel canonical decomposition allowing practitioners to analyze any data valuation method as the combination of two parts: a utility function that captures characteristics from a given model and an aggregation procedure that merges such information.
We also propose to use Gaussian Processes as a means to easily access the utility function on ``sub-models'', which are models trained on a subset of the training set. 
The strength of our approach stems from both its theoretical grounding in Bayesian theory, and its practical reach, by enabling fast estimation of valuations thanks to efficient update formulae.
\end{abstract}

\section{Introduction}
Recently, the field of Machine Learning (ML) has witnessed the development of increasingly complex models and algorithms, due in part to a surge in the amount of available data. 
However, not all data are created equal, and the quality and relevance of the data play a crucial role in the success of any ML model. 
Data Valuation (DV) , the process of quantifying the value of data for a specific task, has emerged as a critical concern~\citep{jia2019towards, pmlr-v97-ghorbani19c, pmlr-v119-ghorbani20a}.
More precisely, the DV field has witnessed a growing interest in measuring the impact of individual data points on the performance of ML algorithms. 

One common approach in this regard is to evaluate how the removal of a specific datum affects the overall model performance.
This DV technique, called \textit{Leave One Out} (LOO), provides a measure of the importance of this data point.
To capture group effects between inputs, a classical improvement of the LOO is the \textit{Data Shapley} approach~\citep{pmlr-v97-ghorbani19c}.
This method finds its theoretical roots in game theory and considers all \textit{coalitions} or groups of variables, comparing performances with or without the specific datum of interest. 
However, a critical limitation of this approach lies in its fundamental requirement to compute the performance of a trained ML algorithm for all possible subsets of the training dataset. 
This requirement, although theoretically informative, often proves to be computationally infeasible due to the exponential growth in the number of subsets as a function of dataset size.

To address this computational challenge, we propose an innovative approach that leverages the strengths of Gaussian Processes (GPs) in the context of DV.
GPs~\citep{matheron1963principles,williams2006gaussian} are a versatile tool in ML known for their ability to model complex relationships in data, to provide uncertainty estimates, to seamlessly incorporate new data thanks to analytical update formulae and to make goal-oriented exploration of the input space. 
In this context, GPs offer a novel solution to the computational bottleneck associated with traditional DV techniques.

In our work, we leverage the unique structure of GPs enables us to perform explicit and immediate computations of sub-models, which are models trained on subsets of the complete dataset.
In the framework of DV, this means \emph{that we can easily and at low cost access information on models trained without a given datum of interest, and compare it with the model trained on all data points}.
Thus, in balance with the exhaustive approaches required by Data Shapley and similar techniques -- even if this can be alleviated by methods such as truncated Data Shapley --, GPs allow us to obtain insights into the impact of individual data points by constructing easily these sub-models.
This is achieved through the GPs' distribution inherent ability to be completely summarized by their mean and covariance function, which can both be expressed as explicit weighting of values obtained for the training set.
By working within the GPs framework, we circumvent a large part of the computational intractability associated with traditional DV methods, making it feasible to efficiently assess the importance of data points in large datasets. 
Thus, our approach not only provides practical advantages but also offers a principled and data-driven way to quantify the value of individual data points, contributing to more informed decision-making in data pre-processing, feature selection and model development. 

The outline of the paper is as follows.
First in Section~\ref{sec_background_dv}, we review both the DV and GP literature and related work.
Then, in Section~\ref{sec_framework}, we propose a canonical decomposition of DV methods as two intertwined elements: a utility function and an aggregation procedure.
We then discuss how the Integrated Variance of GPs can be used as a possible utility function to speed up the computations of DV.
Finally, in Section~\ref{sec:exp}, we demonstrate experimentally its usefulness before concluding in Section~\ref{sec:conclu}.

\section{Background}
\label{sec_background_dv}

\textbf{Notations.} 
We will use the following notations in the rest of the paper. 
We consider a dataset 
$\mathcal{D}$ composed of $n$ entries $\bs{x}_1, \cdots, \bs{x}_n$ that are the realizations of a random variable $\bs{X}$ that belongs to a probability space $(\mathcal{X},\mathcal{T}, \mathbb{P}_{\bs{x}})$. 
Let $f(\bs{x}, \mathcal{D})$ denote a model that takes as input a vector $\bs{x}$ and that is trained using the dataset $\mathcal{D}$.
In the ML literature, the model is often obtained by choosing $f(\bs{x};\mathcal{D}) = \argmin_{g \in \mathcal{G}} L_n(g,\mathcal{D})$ for $\mathcal{G}$ a certain class of functions and $L_n$ an empirical loss. 
Considering a set $A$, let $\mathcal{D}_{\sim A}$ denote the set $\{ i \in \mathcal{D}, i \not \in A\}$, which is the complementary set of $A$ in the complete dataset -- \emph{e.g.} $\mathcal{D}_{\sim i}$ for the complete dataset except for the datum $i$.

\subsection{Related Work on Data Valuation}

At its core, Data Valuation (DV) aims at quantifying the contribution of individual data point—or datum—in training a model, assessed using a test set. 
Originally, the development of DV arises from the increasing desire of individuals and organizations to quantify the value of their data~\citep{jia2019towards}, hence the use of the economic term ``valuation''. 
While Ghorbani and Zou share this perspective, they also regard DV as a method for assessing data quality~\citep{pmlr-v97-ghorbani19c}. 
In particular, DV is useful when training models on noisy datasets, as it helps to identify and remove data points that degrade the model's performance. 
In addition, as ML gains wider adoption, determining the influence (or lack thereof) of individual data points in data-driven decisions has become crucial. 
Besides, DV can be instrumental in performing data removal, which can be defined in the ML context as determining the data points that are absolutely necessary to the performance of a model.
Doing so can lead to lower storage costs, together with reduced privacy risks.

Shapley values, a concept originated from cooperative game theory~\citep{Shapley+1953+307+318}, are often employed as a reference and benchmark for distributing value among data. 
Its strength lies in its fair axioms and the stability of values it provides.
In contrast, naive methods like LOO, in which two models are trained -- one on the whole dataset and one on the dataset except the datum of interest, have been shown to be less reliable~\citep{pmlr-v97-ghorbani19c}, albeit faster to compute. 
Indeed, the major drawback of the Shapley values is that calculating them for all data points has an exponential complexity with respect to the size of the dataset, which can render the computation of data values impractical. 

Thus, since the introduction of DV in ML, much of the effort has been directed towards making it computationally feasible, juggling between the need to explore numerous subsets to capture datapoint interactions and the cost associated with computations on sub-models. 
One strategy is to approximate it using Monte Carlo methods~\citep{pmlr-v97-ghorbani19c}.
Another possibility is to adopt alternative concepts such as the Banzhaf~\citep{wang2022data} or the Beta values~\citep{kwon2021beta}. 
Their associated aggregation procedures can be linked to the game theory literature on semi-values, of which Shapley values are a special case. 
However, one of their advantages is that there are more efficient and robust methods for estimating values.
Nevertheless, these methods still necessitate retraining the model to compute the contributions of data points. 
A third approach involves employing less computationally demanding models or specific families of models, such as nearest neighbors algorithms, which due to their structure do not require computing the contribution of data for every subset~\citep{jia2019efficient}. 
In addition, data value can also be defined by its contribution to the distance between training and test datasets, leading to a learning-agnostic DV framework like LAVA, which is computationally more efficient~\citep{just2023lava}.

\subsection{Primer on Gaussian Processes}


Originally introduced within the domain of geo-sciences~\citep{krige1951statistical,matheron1963principles}, GPs became a popular tool in ML~\citep{williams2006gaussian} due to several advantages. 
First, GPs models are 
flexible non-parametric models that provide a built-in uncertainty quantification, which can be highly valuable for goal-oriented tasks. 
Indeed, they enable modeling of a large class of random functions and work on a variety of inputs: continuous, categorical, structured (\emph{e.g.}, molecules, graphs, etc.).
As such, they are proxies of choice for performing Bayesian optimization~\citep{snoek_practical_2012}, stochastic inversion~\citep{travelletti_uncertainty_2022} and other statistical inference tasks~\citep{marrelCalculationsSobolIndices2009}.
Traditionally, these models are considered particularly efficient in a low data regime and can provide valuable insights in applications in which data collection is costly or difficult. 
While originally criticized for their scalability issues, recent improvements in the implementation of GPs have led to a renewal of their use. 
More precisely, their improved applicability comes from efficient approximation schemes such as Gaussian Markov Random Fields~\citep{lindgren_explicit_2011} and variational approaches \citep{hensman_gaussian_2013}, as well as from improved capacities of the hardware available~\citep{wang_exact_2019} (\emph{i.e.} the use of GPUs rather than CPUs).
Hereafter, we review the basics of GP regression (GPR) from a spatial statistics point of view by presenting the linear unbiased estimators and refer the reader to textbooks on the matter such as~\citet{williams2006gaussian} and references therein for the ML-centric perspective.

\textbf{Basics of GPR.}
A GP is a collection of random variables, any finite number of which have a joint multivariate Gaussian distribution, which makes it a popular choice as a prior over functions. 
A GP $Z$'s distribution is characterized by its mean function: $m(\bs x):=\mathbb{E}[Z(\bs x)]$ and its covariance kernel $k(\bs x, \bs y):= \text{Cov}(Z(\bs x), Z(\bs y))$, for any $\bs x, \bs y \in \mathcal{X}$. 
We represent it by $Z\sim \mathcal{GP}(m, k)$.
Hereafter, our focus is on GPR, a non-parametric regression approach. 
More precisely, we consider a dataset of $n$ pairs of inputs - outputs: $\mathcal{D}=(\bs x_i)_{1\leq i \leq n}$, along with each observed output $z_i \in  \mathbb{R}$ assumed to be a realization of the function to be modeled at $\bs{x_i}$, with $z_i = \mu(\bs{x_i}) + Z(\bs{x_i})$. 
Here, $\mu$ is a trend function modeling the deterministic part of the observation and $Z\sim \mathcal{GP}(0, k)$ is a centred GP capturing the spatial stochastic dependency.
The three primary variants of GPR are: (1) when $\mu$ is fully known (\emph{Simple Kriging}), (2) when $\mu$ is an unknown constant (\emph{Ordinary Kriging}) or (3) when $\mu$ is a linear combination of given basis functions with unknown coefficients (\emph{Universal Kriging})~\citep{omre1989bayesian}. 

In practical scenarios, the trend is generally unknown, which makes Simple Kriging not suitable. 
Since Universal Kriging encompasses Ordinary Kriging as a special case, we will focus on it.
Assuming that the trend is given by $\mu(\bs{x})=\sum\limits_{j=1}^p \beta_j f_j(\bs{x})$, the Universal Kriging predictor at a point $\bs{x} \in \mathcal{X}$ is linear in the observed values $\bs{z} =\left( z_i\right)_{1\leq i\leq n}$:
\begin{equation}
    \hat Z(\bs{x}):= \bs{\lambda}(\bs{x})^\top {z},
\end{equation}
in which the vector of weights $ \bs{\lambda}(\bs{x})$ is given by:
\begin{equation}
\label{eq:KrigingWeights}
    \left(\begin{matrix}
        K & F \\
        F^\top & \bs{0} 
    \end{matrix}\right)^{-1} \left(\begin{matrix}
       \bs{k}(\bs{x})\\
       \bs{f}(\bs{x})
    \end{matrix}\right)=
    \left(\begin{matrix}
       \bs{\lambda}(\bs{x})\\
       \bs{*}
    \end{matrix}\right)
\end{equation}
with $K$ being the matrix $\left(k(\bs{x_i}, \bs{x_j})\right)_{1\leq i, j \leq n}$, $F$ being the matrix $\left(f_i(\bs{x_j}) \right)_{1\leq i \leq p, 1\leq j \leq n}$, $\bs{k}(\bs{x})$ being the vector of $\left( k(\bs{x_i}, \bs{x})\right)_{1\leq i\leq n}$ and $\bs{f}(\bs{x})$ the vector  $\left( f_i(\bs{x})\right)_{1\leq i\leq p}$. 
Therefore, the weight for each observation combines insights from $\bs{f}(\bs{x})$ pertaining to the trend, as well as from $\bs{k}(\bs{x})$ relating to the spatial structure of our model.

Two key features of GP regression are valuable properties for this research.
First, the inclusion of a new observation $(\bs{x_{n+1}, z_{n+1}})$ in $\mathcal{D}$ leads to an updated Kriging weight system, extending the one described in Equation~\ref{eq:KrigingWeights}.
Second, studying the residuals, defined as $Z(\bs x) - \hat Z (\bs x)$, provides valuable insights on the model's accuracy and effectiveness.

\textbf{Update formula.}
The key part of GPR is accessing the inverse of the covariance matrix $K$.
However, when adding new data, this matrix changes and its size increases.
Naively, one would need to invert the new matrix that includes the new points, without any prior information.
This method can be quite expensive, especially when adding few points to an otherwise large dataset -- \emph{e.g.} adding one point to a dataset of 1000 points would lead to the inversion of a $1001\times1001$ matrix regardless of the knowledge on the covariance sub-matrix of size $1000\times1000$.
This issue can be addressed using the Schur complement, leveraging already known information on the covariance matrix to update it efficiently when adding datapoints.
This result can be found in~\citet{ginsbourger2023fast}.

\textbf{Insights on the residuals.}
In our case, we are also interested in the Integrated Variance of the GP predictor. 
The residuals’ variance $Var(Z(\bs{x}) - \hat Z_{\mathcal{D}}(\bs{x}))$ quantifies the uncertainty of the predictor $\hat{Z}_{\mathcal{D}}$ in its prediction at a given point $\bs{x}$.
As more and more datapoints are used for the predictor, predictions become closer to realizations of the model 
and the quantity $Var(Z(\bs{x}) - \hat Z_{\mathcal{D}}(\bs{x}))$ decreases for each $\bs{x}$.
This quantity is directly accessible by computing the inverse of the matrix 
\begin{equation}\label{eq:cov_matrix}
    M_\mathcal{D}(\bs{x}) = \begin{pmatrix}
        k(\bs{x},\bs{x}) & k(\bs{x},\bs{x}_{\mathcal{D}}) & F(\bs{x})\\
        \bs{k}(x) & K & F \\
        F(\bs{x})^T & F & 0 \\
    \end{pmatrix}.
\end{equation}
Note that, again, we use the covariance matrix $K$ whose inverse is readily accessible and whose update can be done following the previous point.
We are also interested in the global uncertainty of the predictor.
This is quantified by the Integrated Variance, which can be understood as the aggregated remaining uncertainty of the predictor:
\begin{equation}
    IV(\mathcal{D}):= \int Var(Z - \hat Z_{\mathcal{D}}) d\mathbb{P}_\mathbb{X}. 
\end{equation}

This can be expressed in terms of coefficients of $M_\mathcal{D}^{-1}(\bs x) $, by denoting $M_D^{-1}(\bs x)[1,1]$ the top left coefficient and taking:

\begin{equation}
    IV(\mathcal{D}) = \int_{\bs x \in \mathcal{X}} M_\mathcal{D}^{-1}(\bs x)[1,1] d\mathbb{P}_\mathbb{X}(\bs x).
\end{equation}
However, inverting the matrix $M_D(\bs x)$ for any given input point $\bs x$ can be costly. 
This issue can be, once again, solved by using the Schur complement of the covariance matrix. 

\begin{theorem}[Iterative update of Integrated Variance]
\label{thm:updateGP}
For a set $A \subset \mathcal{D}$, let $\Tilde{K}_{A} := M_{A}[\sim 1,\sim 1]$ the matrix obtained by deletion of first row and column of $M_A$ and the covariance matrix of the data point included in $A$. 
When adding a datum $i$, the inverse of the matrix $\Tilde{K}_{A\cup \{i\}}$ can be obtained using the Schur complement formula knowing only the inverse of $K_{A}$ -- see~\citet{ginsbourger2023fast} or explicit formula in Appendix~\ref{app:schur}.
Moreover, let $IV(A)$ be the integrated variance of the GP from the dataset $A \subset \mathcal{D}$.Then we have an explicit formula for $IV(A \cup \{i\})$ that we provide in Equation~\ref{eq:endgameupdate} in Appendix~\ref{app:schur}.
\end{theorem}

More precisely, one can see two different uses of the Schur complement here: one to update the covariance matrix and one to extract the quantity $Var(Z - \hat Z_{\mathcal{D}})$.


\section{Proposed Data Valuation Framework}
\label{sec_framework}

In this section, we describe our proposed framework for showing that most, if not all, DV tools can be decomposed in two parts that we describe in the following subsection.
This dichotomy allows to conduct a generic and systemic analysis of DV tools while providing the practitioner with additional insights and knowledge on what is actually captured by these indices.
Then, we provide several examples of different DV methods, comparing the differences between them, along with some hints on classical limitations that they have when confronted with real world datasets (\emph{e.g.}, the need for sub-models).

\subsection{Utility Functions and Aggregations}

As discussed previously, there are various DV indices that have been proposed in the literature,  
 but as we detail hereafter the vast majority 
of them can be decomposed as two elements.

\textbf{A utility function.} This corresponds to a way of measuring some characteristic of a model, which is usually done through a function $\Phi$ quantifying the model performance. 
Common choices for this performance measure include the accuracy, the RMSE, the F1-score, the area under the ROC curve (AUC), etc.
However, some indices can be simpler (\emph{e.g.}, the model output at a given point can be used directly) or more complex (\emph{e.g.}, a composite metric of the fairness of the model).
Similarly, this utility function can be either computed using a test set -- for instance, after an initial split of the data done before the DV procedure (\emph{e.g.}, the accuracy of the model with respect to the test set) -- or using exact computation of characteristics (\emph{e.g.}, the expectation of a given functional of the model such as a fairness criterion). 
We emphasize that in this canonical decomposition, a usual train-test split of the data as commonly used in the literature is done before this procedure, with train data (\emph{i.e.}, $\mathcal{D}$ used for model training for obtaining $f(\bs{x};\mathcal{D})$) -- and test data used for computing later quantities -- such as $\Phi[f(\cdot;\mathcal{D})]$ if such a utility function needs empirical data. 
The choice of the utility function is the most flexible element of DV and usually the part in which practitioners can use expert knowledge the most easily to choose this function.
    
Note that in some situations, this utility function is only computed on some specific subsets of the complete dataset $\mathcal{D}$ according to the needs of the aggregation part (which we discuss afterwards) (\emph{e.g.}, $\Phi[f(\bs{x};\mathcal{D})]$ and $\Phi[f(\bs{x};\mathcal{D}_{\sim i})]$ for any datum $i$). 
However, in general settings, this quantity needs to be computed for all models trained from subsets of the dataset of $\mathcal{D}$. 
More precisely, generally we need to be able to compute $\Phi[f(\bs{x};A)]$ for any $A \subset \mathcal{D}$, which can be expensive and is our motivation for GP usage.

\textbf{An aggregation procedure.} 
Once we have the value of the utility function on all possible subsets of $\mathcal{D}$, we need to compare them depending on the presence or absence of the datum $i$ to identify its influence on the training of the model.
The simplest of these aggregation procedures is the LOO described previously, in the sense that the valuation of the datum is the direct comparison between performances of these two models. 
However, by doing this, we are limited to direct effects of the datum’s presence in the dataset. 
Various refinements have been proposed to take into account indirect effects -- that is belonging of the datum to specific subsets. 
Among them, one can think about Shapley values~\citep{Shapley+1953+307+318} and recent variations such as Banzhaf~\citep{wang2022data} or Beta Shapley values~\citep{kwon2021beta} found in DV literature. 
Nonetheless, other aggregation values can be of interest, including more general semi-values and other concepts from cooperative game theory.
Note that the aggregation procedure indicates how many sub-models are needed for accessing DV indices.
For instance, simple aggregation procedures may not be very informative or may only provide first-order insights but can be computed with very few sub-models.

\begin{figure}[h!]
    \centering
    \hspace*{-1cm}
\begin{tikzpicture}[remember picture, scale=0.975]

    \node[nodeStyle] (A) at (0,0) {Impact of a\\ datum in $D$};
    \node[nodeStyle] (B) at (3.5,0) {Measured through $\Phi$\\on sub-models};
    \node[nodeStyle] (C) at (7,0) {Aggregated for\\Data Valuation};

    \node[font=\footnotesize, anchor=north, align=center] (B2) at ($(B.south)+(0,-0.5cm)$) {Trained on singleton\\
    $\Phi\left[ f(\bs x; \{\bs{\tikzmarknode{highlighted1}{\bs{x_1}}}\}) \right]$};
    
    \node[font=\footnotesize, anchor=north, align=center] (B3) at ($(B2.south)+(0,-0.2cm)$) {Trained on pairs containing $\bs{x_1}$\\
    $\Phi\left[ f(\bs x; \{\tikzmarknode{highlighted2}{\bs{x_1}}, \tikzmarknode{highlighted21}{\bs{x_2}}\}) \right]$ \\ $\cdots$ \\ $\Phi\left[ f(\bs x; \{\tikzmarknode{highlighted3}{\bs{x_1}}, \bs{x_n}\}) \right] $};
    
    \node[font=\footnotesize, anchor=north, align=center] (B4) at ($(B3.south)+(0,+0.25cm)$) {\vdots};
    
    \node[font=\footnotesize, anchor=north, align=center] (B5) at ($(B4.south)+(0,+0.05cm)$) {Trained on everything\\
    $\Phi\left[ f(\bs x; \{\tikzmarknode{highlighted4}{\bs{x_1}},  \tikzmarknode{highlighted22}{\bs{x_2}}, \dots, \bs{x_n}\}) \right]$};

    \node[font=\footnotesize, anchor=north, align=center] (A2) at ($(A.south)+(0,-0.5cm)$) {For $\bs{x_1}$};
    \DataTable{0}{-2.6}{0.35}{blue}{};
    \coordinate (StartBrace) at ($(B2.north)+(2.8,0)$);
    \coordinate (EndBrace) at ($(B5.south)+(2.8,0)$);
    \coordinate (MiddleBrace) at ($(StartBrace)!0.5!(EndBrace)$);

    \draw[decorate, decoration={brace, amplitude=15pt, mirror}] 
   (EndBrace) -- (StartBrace);
    \coordinate (DV1Coord) at ($(B3.west)!(C.south)!(B3.east)+ (+0.5cm,-0.15cm)$);
    \node[circle, draw, anchor=center] at (DV1Coord) {$DV_1$};

    \draw[line width=0.1pt] (-0.5,-0.675) -- (8.,-0.675);  \draw[line width=0.1pt] (-0.5,-5.5) -- (8.,-5.5);   \draw[line width=0.1pt] (-0.5,-10.60) -- (8.,-10.60); 
    \node[nodeStyle] (D) at (0, -5.5)  {\\};
    \node[font=\footnotesize, anchor=north, align=center] (D2) at ($(D.south)+(0,-0.0cm)$) {For $\bs{x_2}$};
    \node[nodeStyle] (E) at (3.5, -5.5) {\\};
    \node[nodeStyle] (F) at (7, -5.5) {\\};
    \node[font=\footnotesize, anchor=north, align=center] (E2) at ($(E.south)+(0,0cm)$) {Trained on singleton\\
    $\Phi\left[ f(\bs x; \{ \tikzmarknode{highlighted23}{\bs{x_2}}\}) \right]$};
    \node[font=\footnotesize, anchor=north, align=center] (E3) at ($(E2.south)+(0,-0.2cm)$) {Trained on pairs containing $\bs{x_2}$\\
    $\Phi\left[ f(\bs x; \{\tikzmarknode{highlighted5}{\bs{x_1}},  \tikzmarknode{highlighted24}{\bs{x_2}}\}) \right] $\vspace*{0.05cm}\\  $\cdots$\\ $\Phi\left[ f(\bs x; \{ \tikzmarknode{highlighted25}{\bs{x_2}}, \bs{x_n}\}) \right] $};
    \node[font=\footnotesize, anchor=north, align=center] (E4) at ($(E3.south)+(0,+0.25cm)$) {\vdots};
    \node[font=\footnotesize, anchor=north, align=center] (E5) at ($(E4.south)+(0,+0.05cm)$) {Trained on everything\\
    $\Phi\left[ f(\bs x; \{\tikzmarknode{highlighted6}{\bs{x_1}},  \tikzmarknode{highlighted26}{\bs{x_2}}, \dots, \bs{x_n}\}) \right]$};
    
    \DataTable{0}{-7.5}{0.35}{}{orange};
    \coordinate (StartBrace2) at ($(E2.north)+(2.8,0)$);
    \coordinate (EndBrace2) at ($(E5.south)+(2.8,0)$);
    \coordinate (MiddleBrace2) at ($(StartBrace2)!0.5!(EndBrace2)$);

    \draw[decorate, decoration={brace, amplitude=15pt, mirror}] 
   (EndBrace2) -- (StartBrace2);
    \coordinate (DV2Coord) at ($(E3.west)!(C.south)!(E3.east)+ (0.5cm,-0.15cm)$);
    \node[circle, draw, anchor=center] at (DV2Coord) {$DV_2$};

    \draw[->, color=gray] ($(E3.south west)+(1cm,0.85cm)$)  |- ++(-1.2,0) |- ($(B3.west)+(0.75,0.2cm)$);
    \node[font=\tiny, text=gray, anchor=north, align=center] at ($(E3.south)+(0cm,+1.05cm)$) {Already seen with $\bs{x_1}$};
    
    \draw[->, color=gray] ($(E5.south west)+(0.9cm, 0.0cm)$) |- ++(-1.35,0) |- ($(B5.west)+(0.05,-0.125cm)$);
    \node[font=\tiny, text=gray, anchor=north, align=center] at ($(E5.south)+(0cm,+0.2cm)$) {Already seen with $\bs{x_1}$};
    \begin{scope}[on background layer]
        \fill[blue, opacity=0.2] ($(highlighted1.south west)+(-0.01,-0.05)$) rectangle ($(highlighted1.north east)+(0.01,0.1)$);
        \fill[blue, opacity=0.2] ($(highlighted2.south west)+(-0.01,-0.05)$) rectangle ($(highlighted2.north east)+(0.01,0.1)$);
        \fill[blue, opacity=0.2] ($(highlighted3.south west)+(-0.01,-0.05)$) rectangle ($(highlighted3.north east)+(0.01,0.1)$);
        \fill[blue, opacity=0.2] ($(highlighted4.south west)+(-0.01,-0.05)$) rectangle ($(highlighted4.north east)+(0.01,0.1)$);
        \fill[blue, opacity=0.2] ($(highlighted5.south west)+(-0.01,-0.05)$) rectangle ($(highlighted5.north east)+(0.01,0.1)$);
        \fill[blue, opacity=0.2] ($(highlighted6.south west)+(-0.01,-0.05)$) rectangle ($(highlighted6.north east)+(0.01,0.1)$);
        
        \fill[orange, opacity=0.2] ($(highlighted21.south west)+(-0.01,-0.05)$) rectangle ($(highlighted21.north east)+(0.01,0.1)$);
        \fill[orange, opacity=0.2] ($(highlighted22.south west)+(-0.01,-0.05)$) rectangle ($(highlighted22.north east)+(0.01,0.1)$);
        \fill[orange, opacity=0.2] ($(highlighted23.south west)+(-0.01,-0.05)$) rectangle ($(highlighted23.north east)+(0.01,0.1)$);
        \fill[orange, opacity=0.2] ($(highlighted24.south west)+(-0.01,-0.05)$) rectangle ($(highlighted24.north east)+(0.01,0.1)$);
        \fill[orange, opacity=0.2] ($(highlighted25.south west)+(-0.01,-0.05)$) rectangle ($(highlighted25.north east)+(0.01,0.1)$);
        \fill[orange, opacity=0.2] ($(highlighted26.south west)+(-0.01,-0.05)$) rectangle ($(highlighted26.north east)+(0.01,0.1)$);
    \end{scope}
\end{tikzpicture}
\caption[Pipeline DV]{Generic pipeline for Data Valuation.}
\label{fig:pipelineDV}
\end{figure}
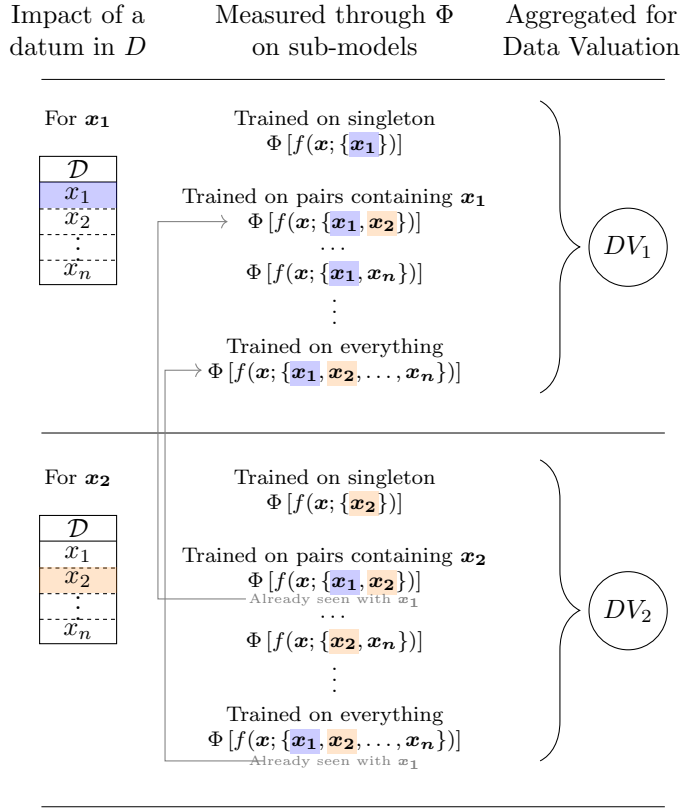

To formalize this framework, we first introduce the following definition.
\begin{definition}[General data valuation of a set of indices]
Let $\Phi: \mathbb{L}^2(\mathbb{P}_{\bs{X}}) \mapsto \mathbb{R}$ be a utility function and $Agg_i:\mathcal{P}(\mathcal{D}) \mapsto \mathbb{R}$ be an aggregation procedure for the datum $i$. 
Then a \textit{general Data Valuation set of indices} $\{DV_{i}(\Phi, Agg_i, f,\mathcal{D})\}$ is defined by its elements 
    \begin{equation}
        DV_i(\Phi, Agg_i, f,\mathcal{D}) := Agg_i(\{\Phi(f(\bs{x};A)), A \subset \mathcal{D}\}).
    \end{equation}
\end{definition}
Note that the function $Agg_i$ depends on the datum of interest but can be applied to all the quantities $\Phi[f(\bs{x};A)], A \subset \mathcal{D}$.
Figure~\ref{fig:pipelineDV} also illustrates the envisioned pipeline with an example in which the aggregation procedure includes all coalitions to which the datum $i$ belongs.

The main difficulty for performing DV is to access the quantity $\Phi[f(\bs{x});A]$ for some subset $A$ of $\mathcal{D}$ when working with meaningful utility functions.
In the general case, the training is not done for all subsets of the data and the intermediary models and values are not kept. 
However, in this paper we chose to focus on the GPs as a means to alleviate this problem.

\subsection{Examples of DV indices}
Hereafter, we describe some examples of DV indices that can be used in practice. 
Following our proposed framework, these indices can be interpreted as a specific choice of utility function and aggregation procedure, thus allowing them to capture specific characteristics of the dataset. 

\textbf{Naive utility function, naive aggregation.} 
The first example is the classical DV procedure that arises in the classification framework.
In that case, we have a random variable $Y$ that is predicted from covariates $\mathbf{X}$ thanks to the model $f$.
The utility function in the DV procedure is the accuracy of the trained model and the aggregation procedure simply computes the difference between the utility function on the model trained with the complete dataset and the utility function taken on the model trained with the truncated dataset without the datum of interest. 
More precisely, we have:
\begin{align}
    \Phi[f(\bs{x};A)] & = \int \mathbbm{1}_{f(\bs{x};A) = Y} d\mathbb{P}_{\bs{x}}(\bs{x}) ;\\ Agg_i(\{\Phi[f(\bs{x};A)];A \subset \mathcal{D}\}) & = \Phi[f(\bs{x};\mathcal{D}] - \Phi[f(\bs{x};\mathcal{D}_{\sim i}].
\end{align}

In terms of benefits, this set DV indices is simpler to interpret and compare than more complex set of indices.
It also needs only $n+1$ model training to obtain all valuations, namely the utility function on the complete dataset and $n$ utility functions on truncated datasets. 
However, one of the biggest drawbacks of this technique is that it quantifies only ``first-order'' importance of a datum -- which means that the datum only provides the information of its contribution when added to an almost complete dataset. 
Other more subtle effect -- \emph{e.g.} when a datum is interesting for the complementary information it provides when linked with specific subsets of data -- cannot be captured by such indices.
Additionally, another issue is that the valuations obtained do not directly sum to one and have no reason to be positive as additional data may deteriorate the accuracy of the model.

\textbf{Data Shapley}
\label{ex:Shapley_Game_Theory}
Data Shapley indices~\citep{pmlr-v97-ghorbani19c} are one of the most used DV techniques in the literature. 
The utility function for this set of indices can be readily changed according to the use case and is not the core of this valuation. 
However, the aggregation procedure originated from game theory and can be linked with several other fields such as explainability.
Here, the objective is to go beyond the ``first-order'' aggregation explained in the previous example by taking into account the importance of the datum of interest not only in the complete dataset but also in every possible subsets (or coalitions) of this dataset. 
To match a certain number of axioms defined by \citet{Shapley+1953+307+318}, the weight of the utility functions are uniquely defined and allow to obtain a set of data valuations summing to one.
Additionally, the valuations obtained can be easily interpreted as the coalitional importance of each datum in the training dataset.
The aggregation procedure is given by the following formula: 
\begin{equation}\label{def:DS}
        DS_i(\Phi,f,\mathcal{D}) := (n!)^{-1} \sum_{S \in \mathcal{D}_{\sim i}} \Phi[f(\bs{x}; S\cup\{i\})] -  \Phi[f(\bs{x}; S].
\end{equation}
Note that this definition is equivalent to other definitions that focus on coalitions enumeration -- given for instance in the literature~\citep{pmlr-v119-ghorbani20a} -- by considering all possible permutations, a classical alternative vision for Shapley values.
Thanks to this formulation, it allows for an easier understanding of the Monte-Carlo estimator of this aggregation procedure as this estimator is a direct empirical plug-in in this equation.
The most problematic part of these semi-values based indices is the need to evaluate the utility function on all possible coalitions -- that is to say, on all $2^n$ possible trained models. 
As this is usually computationally unfeasible, several estimations have been proposed to alleviate this issue~\citep{pmlr-v119-ghorbani20a}. 
Later, we will use the Monte Carlo estimator found in particular in~\citet{pmlr-v97-ghorbani19c} for DV or~\citet{da2021basics} in Explainability. 

\textbf{Integrated variance and Gaussian processes.}

The final example introduced here is the novelty of our approach.
Its aim is to give a utility function to the practitioner -- more precisely the Integrated Variance of the Gaussian Process -- that is readily understandable. 
Such a quantity is rooted in Bayesian theory as a measure of uncertainty of the trained model and its training dataset, compared with the true phenomenon modeled. 
While the aggregation procedure could easily be chosen as the naive one, we consider the more challenging and informative aggregation function of the DataShapley indices.
One of the main issue, as evoked above, is the high number of trained models needed. 
While this is still true when using Gaussian Processes, we can obtain analytical expressions for the Integrated Variance and its updates when adding a datum.
This is what makes the computation of the utility function faster, since we do not have to compute everything from scratch each time we consider a new coalition.
Furthermore by clever usage of the Schur complement -- detailed in Appendix~\ref{app:schur}, see~\citet{zhang2006schur, gallier2010notes}  for in-depth details --, we propose an algorithm to obtain results that are computationally realistic and allow for better estimation due to the possibility of more calls to the model in the Monte Carlo loop of the estimator, since we do not spend as much compute for the utility function. 
We emphasize that this framework does not speed-up the aggregation procedure but rather accelerate the computation of the utility function, allowing to spend a greater computational budget in the aggregation procedure (\emph{e.g.}, more runs in the Monte-Carlo loop).

\begin{algorithm}
\caption{Integrated Variance and Shapley Aggregation estimator}\label{alg:IV_ShapAgg}
\KwData{Training set $\mathcal{D}$, full kernel matrix $M_{\mathcal{D}}$, a maximal budget $b_{max}$, a tolerance threshold $\varepsilon$}
\KwResult{Data valuation of training points $DV = \{DV_i, i \in \{1,\cdots,n\}\}$}
Initialize $DV_i \gets 0$ for all $i \in \{1,\cdots,n\}$\;
Initialize $b \gets 0$\;

\While{$b \leq b_{max}$}{
    Initialize $\text{Upd}_i \gets 0$ for all $\{ i \in \{1,\cdots,n\}\}$\;
    Initialize $\text{IV}_{\pi, \emptyset} \gets IV(\emptyset)$\;
    Update the budget: $b \gets b+1$\;
    Draw a random permutation $\pi$ of data points\;
    \For{$j \in \{1,\cdots,n\}$}{
        Use $\text{IV}_{\pi, j-1}:=IV(\{\pi[1],\cdots,\pi[j-1]\})$ to compute $\text{IV}_{\pi, j}:=IV(\{\pi[1],\cdots,\pi[j]\})$ with Equation~\ref{eq:endgameupdate} from Appendix~\ref{app:schur}\;
        Update $\text{Upd}_j \gets \frac{1}{n!}\left(\text{IV}_{\pi, j} - \text{IV}_{\pi, j-1}\right)$\;
    }
    Update $DV_i \gets \frac{b-1}{b} DV_i + \frac{1}{b} \text{Upd}_i$ for all $\{ i \in \{1,\cdots,n\}\}$\;
}
\end{algorithm}

Note that a final benefit of using GPs is that it is versatile while proposing a unified framework for working with tabular and non-tabular data. 
Indeed, depending on the type of data, the only part that needs to be adapted for the GP to work is the kernel. 
This subject has been studied in the literature in recent years, especially when considering GP modeling in biochemistry~\citep{tanimoto1958elementary, tripp2023tanimoto}.
This can be interesting for broader applications of DV, but we leave it as future works.

\section{Experiments}
\label{sec:exp}
In our exploration of DV using GPs, we focus on synthetic data 
and the Boston Housing dataset~\citep{harrison_hedonic_1978}, a widely recognized benchmark in regression tasks. 
We also refer the reader to Appendix~\ref{appendix:experiments} for additional experiments on Boston Housing and for another regression task performed on a variant of the Adult Income Dataset generated through the Folktables framework~\citep{ding2021retiring}.

\subsection{Synthetic data}
\label{subsec:synthexp}
\begin{figure}[h!]
    \centering
    \begin{subfigure}{0.495\linewidth}
        \centering
        \includegraphics[width =0.99\linewidth]{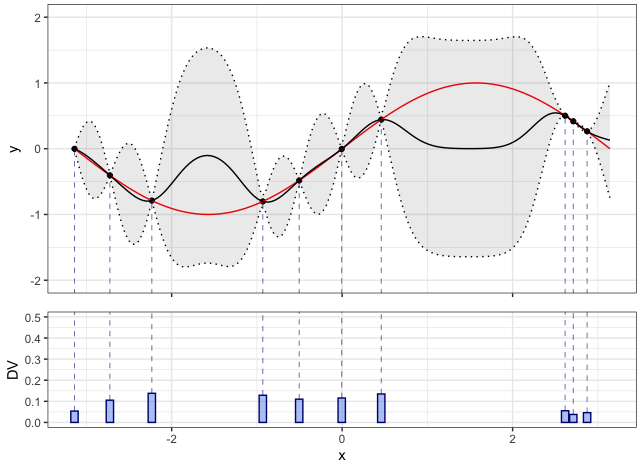}
        \caption{GP with 10 datapoints.}
    \end{subfigure}  
    \begin{subfigure}{0.495\linewidth}
        \centering\includegraphics[width =0.99\linewidth]{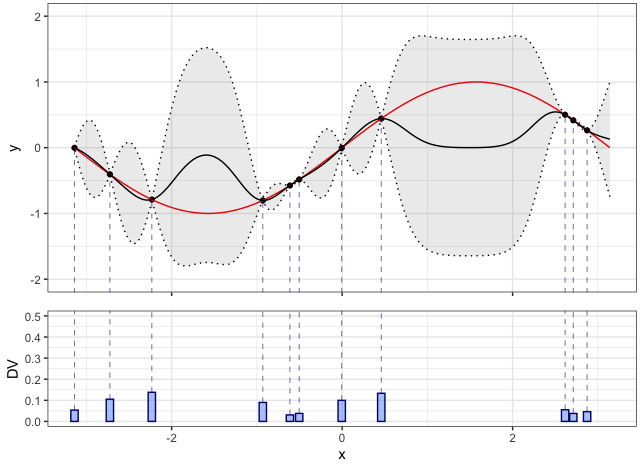}
        \caption{GP with 11 datapoints.}
    \end{subfigure}\\
    \begin{subfigure}{0.495\linewidth}
        \centering\includegraphics[width =0.99\linewidth]{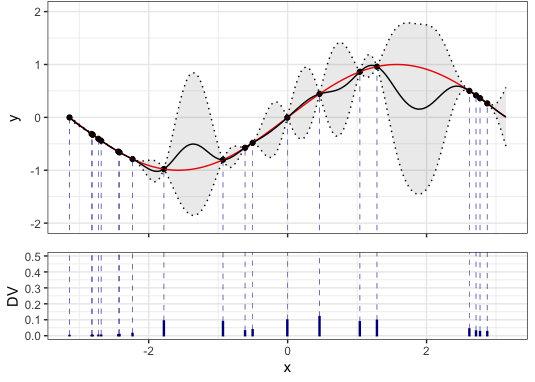}
        \caption{GP with 20 datapoints.}
    \end{subfigure}%
    \caption[GP data assimilation]{Top panels represent the data assimilation of a Gaussian Process. 
    The true model is in red, the predictor in black and the point-wise $95\%$ confidence intervals are represented by the grey ribbon. 
    Bottom panels plot the DV of each datum, with the procedure described in subsection~\ref{subsec:synthexp}.}
    \label{fig:GPiterative}
\end{figure}

We consider the scenario in which data points have been used to train a model $f$ represented in Figure~\ref{fig:GPiterative} as the red curve -- here, a sinus function. 
Following the procedure described in the previous section, we use a GP with a Gaussian kernel -- $k(x,x^\prime)  \propto  \exp(-(x-x^\prime)^2/2)$. 
We perform GPR and at each step, we leverage Schur complement to integrate new datapoints.
We 
compute the DV with IV as a utility function. The results are showcased in Figure~\ref{fig:GPiterative}, where we visualise the evolution of the GPR predictor, the uncertainty reduction when adding new points, and the DV indices.

As expected, we can observe that the points with the biggest importance are isolated points. 
This is due to the fact that removal of an isolated point significantly increases the uncertainty of the neighbouring area. 
Conversely, we can see that point clusters usually contains individuals with low valuation since their contribution is similar one to another. 
More precisely as a whole, the cluster can be highly influential but this influence will be split between several individuals.

\subsection{Boston Housing}

The Boston Housing dataset consists of 506 records with 14 features each, with the objective of predicting the median value of homes in Boston suburbs. 
This dataset was chosen for its relevance to regression tasks and its comprehensive range of features, such as crime rate and property tax rate. 
We perform GPR on noisy observations with a Matérn kernel, reflecting common practices in the literature.
The conducted experiments are Leave-One-Out (LOO), LOO with Schur complement, Data Shapley Value computed without Schur (DSV) and DSV with Schur complement, aiming to evaluate individual data contributions and improve computational efficiency. 
The Schur complement's integration is pivotal in managing the covariance matrix's dynamic nature, especially during DSV calculations, balancing computational stability with efficiency (Appendix~\ref{appendix:experiments}).

\begin{figure}[h!]
    \centering
    \includegraphics[width=0.85\textwidth]{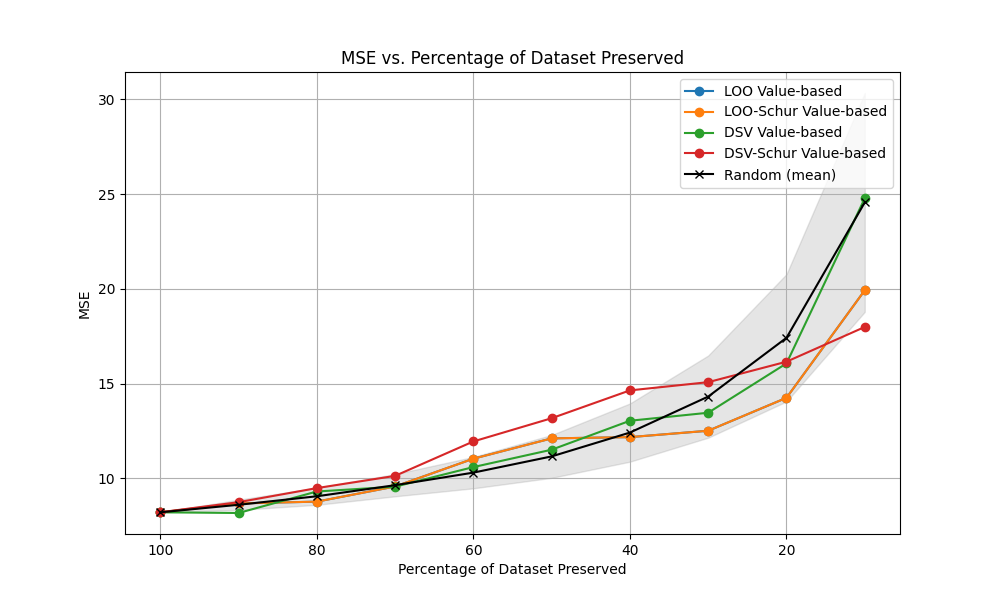} 
    \caption{Comparison of MSE as a function of the percentage of the Boston Housing dataset preserved. 
    The graph illustrates the performance of data valuation methods—LOO, LOO-Schur, DSV, and DSV-Schur against the baseline of random data point removal (mean). Note that LOO and LOO-Schur are equal.
    The MSE increases for all methods as more data is removed, but the valuation-based methods consistently outperform random removal, with a notable advantage as less of 20\% is retained.}

    \label{fig:bh_dataminization} 
\end{figure}

Our analysis for the Boston Housing dataset highlights the importance of using the Schur complement, thus maintaining efficiency without compromising valuation integrity. 
The sanity of our approaches is demonstrated by the consistent results throughout all methods. 
Our implementation yields the same rankings and LOO values no matter the approach considered, as confirmed by a perfect Spearman's coefficient. 
Additionally, a data removal experiment demonstrates 
that DSV-based strategies significantly outperform
random data removal, especially when exceeding $50\%$ dataset reduction, highlighting the effectiveness of data valuation in scenarii of substantial data limitation (\emph{cf.}, Figure~\ref{fig:bh_dataminization}). 
The appendix reveals intricate details concerning the parametrization of the kernel, and computational strategies. 
In particular, we show that 
varying kernel noise levels can be 
used for smoothing the model's output, thus controlling the model's sensitivity to individual data points. 
We also discuss the role of a covariance reset strategy in enhancing the computational stability of DSV valuations with Schur complement (Appendix~\ref{appendix:boston_housing}). 
Finally, in our current benchmark we notice disparities between Data Shapley values when using Schur complement, highlighting that strategic adjustments in the calculation can lead to significant variations in data valuation.

\section{Conclusion and Future Works}\label{sec:conclu}

Our contribution in this paper is twofold. First,   
we have introduced a unifying framework of DV through a canonical decomposition of DV as the combination of two elements: a utility function that extracts some informative characteristics from a model and an aggregation procedure that merges information from values taken by the utility function on all models trained on a subset of the dataset. 
Second, 
we tackled the issue of combinatorial explosion when accessing such models trained on subsets, and have proposed to use of GPs as a means to reduce costs by having readily tractable ``sub-models''. Our careful handling of the algebra in GPR yields low cost estimations of the data valuations.
We focused on one setting where these two axis synergize, by detailing the special case of Integrated Variance as a utility function combined with the Shapley aggregation procedure. We demonstrated the applicability of this approach on synthetic data as well as two others datasets.

These results pave the road for further developments. First, we believe that our canonical decomposition may be a key component for further systematic analysis of DV indices. 
Second, update formulae are a versatile tool that can be used to measure the influence of a coalition, in addition to that of a single datum. 
Finally, GPs usage open a novel and extremely flexible framework for DV, and allows for working with non-tabular data, thus enabling DV in complex dataset.

\bibliography{references}

\newpage
\appendix
\onecolumn

\appendix

\section{Schur decomposition}\label{app:schur}
In this appendix, we provide more details on the Schur complement, along with the proof of Theorem \ref{thm:updateGP}.

First, the Schur complement states the following: for a given invertible matrix we have

\begin{equation}
    \begin{pmatrix}
    A & B\\
    C & D\\
    \end{pmatrix}^{-1} = \begin{pmatrix}
        (A - BD^{-1}C)^{-1} & -(A - BD^{-1}C)^{-1}BD^{}-1 \\
        -D^{-1}C(A - BD^{-1}C)^{-1} & D^{-1} - D^{-1}C(A - BD^{-1}C)^{-1}BD^{-1} \\
    \end{pmatrix}.
\end{equation}

In the context of the covariance matrix of a Gaussian Process, this means that when adding a point $i$ to a set $A$, we have:
\begin{equation}
    K_{A \cup \{i\}}^{-1} = \begin{pmatrix}
        k(x_i,x_i) & k(x_i,x_A) & F(x_i)^T\\
        k(x_A,x_i) & k(x_A,x_A) & F(x_A)^T\\
        F(x_i) & F(x_A) & 0 \\
    \end{pmatrix}^{-1} = \begin{pmatrix}
        \Delta_1 & \Delta_2\\
        \Delta_3 & \Delta_4\\
    \end{pmatrix}\end{equation}
    
    with \begin{multline}\Delta_1 = 
        \left(k(x_i,x_i) -  \left(
            k(x_i,x_A)  F(x_i)^T \right)K_A^{-1} \left(k(x_A,x_i)  F(x_i)\right)^T\right)^{-1},\\
            \Delta_2 = -(k(x_i,x_i) -  \left(
            k(x_i,x_A)  F(x_i)^T \right)K_A^{-1} \left(k(x_A,x_i)  F(x_i)\right)^T)^{-1} \left(k(x_i,x_A)  F(x_i)^T\right)K_A^{-1}, \\
            \Delta_3 = -K_A^{-1}\left(k(x_A,x_i)  F(x_i)\right)^T\left(k(x_i,x_i) -  \left(
            k(x_i,x_A)  F(x_i)^T \right)K_A^{-1} \left(k(x_A,x_i)  F(x_i)\right)^T\right)^{-1}, \\\Delta_4 = K_A^{-1} - K_A^{-1}\left(k(x_A,x_i)  F(x_i)\right)^T\left(k(x_i,x_i) -  \left(
            k(x_i,x_A)  F(x_i)^T \right)K_A^{-1} \left(k(x_A,x_i)  F(x_i)\right)^T\right)^{-1}\\\left(
            k(x_i,x_A)  F(x_i)^T \right)K_A^{-1}. \\
\end{multline}
Note that this only inverse in these formulae is the inverse of the lesser covariance matrix denoted as $K_A^{-1}$, which is known from the previous step in our iteration.

Now, when working with the Integrated Variance, we are interested in the top-left term of the inverse of the matrix $M_A(x)$. Using the Schur decomposition on the matrix $M_A(x)$, one readily obtains that 
\begin{equation}\label{eq:endgameupdate}
    M_{A\cup i}(x)^{-1}[1,1]  = \left(k(x,x) - \left( k(x,x_i) k(x,x_A) F(x)^T\right) K_{A \cup i}^{-1} \left( k(x_i,x) k(x_A,x) F(x)^T\right)^{T}\right)^{-1}.
\end{equation}
This quantity is readily available from the previous development and can be further expanded in lesser terms by developing the products.

\section{Experiments}
\label{appendix:experiments}

This appendix provides additional details regarding the experimental setup.

\subsection{Boston Housing}
\label{appendix:boston_housing}

The Truncated DSV algorithm was configured to perform 1500 iterations for the Boston Housing dataset, with a burn-in period of 50 indices per iteration. The burn-in parameter allows the DSV computation to stabilize, aligning closer to Leave-One-Out (LOO) results.

\begin{figure}[h]
    \centering
    \includegraphics[width=0.8\textwidth]{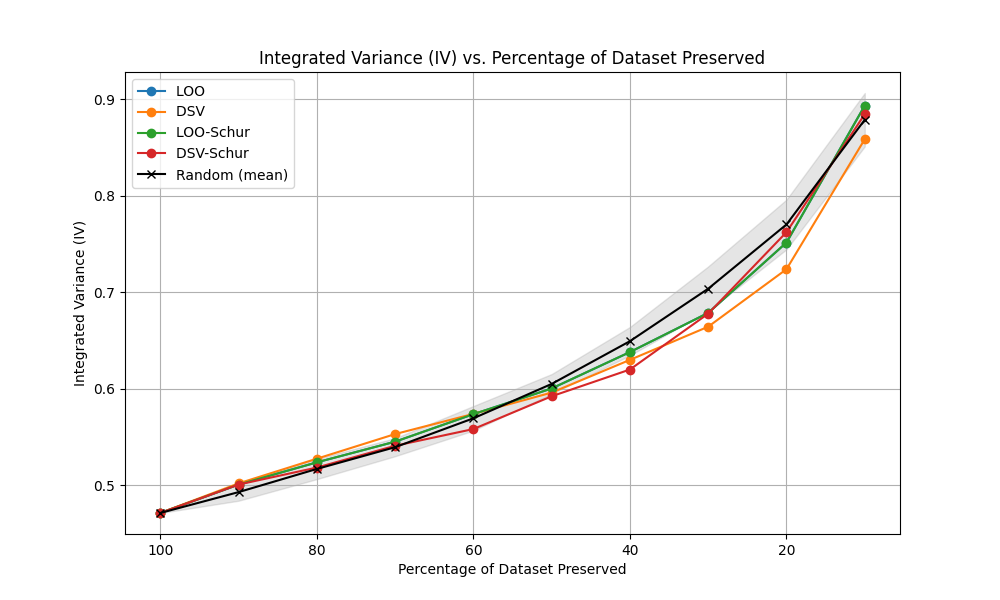}
    \caption{Data valuation impact on Integrated Variance with a kernel noise level of 0.38.}
    \label{fig:iv_high_noise}
\end{figure}

\begin{figure}[h]
    \centering
    \includegraphics[width=0.8\textwidth]{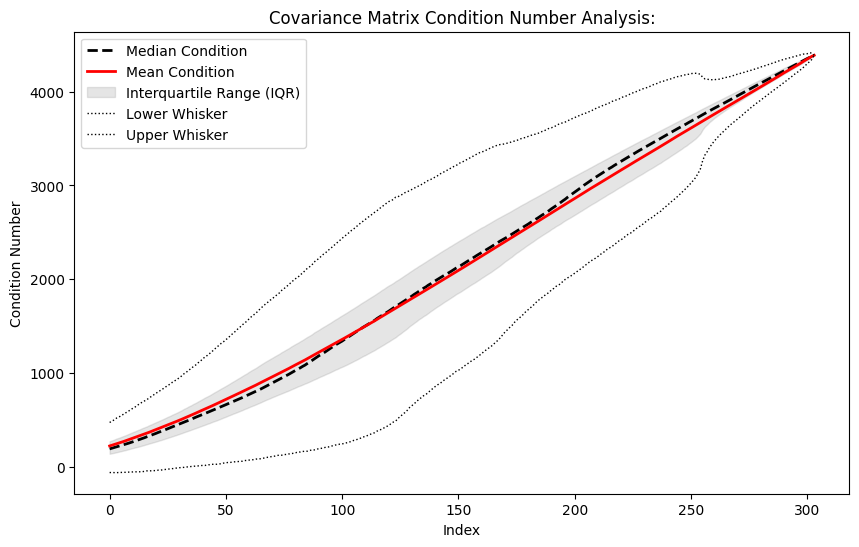}
    \caption{Computational stability analysis using the condition number of the covariance matrix.}
    \label{fig:bh_cov_condition}
\end{figure}

We observed that incorporating noise into the kernel contributes to computational stability at the cost of confidence, as evidenced by a wider Integrated Variance (IV) range when we remove points as shown in Figure~\ref{fig:iv_high_noise}. 
The influence of individual data points on IV becomes more discernible with increased noise levels in the kernel. 
The computational stability was studied by observing the condition number of the covariance matrix (\emph{cf.}, Figure~\ref{fig:bh_cov_condition}). 
It is important to note that adding noise to the kernel allows to control stability but also affects utility.

We also implemented a covariance resetting strategy for instances where the absolute contribution to IV was too large because of computational instability. 
This technique allowed the computations using the Schur complement to align more closely with the DSV values.

\begin{figure}[h]
    \centering
    \includegraphics[width=0.8\textwidth]{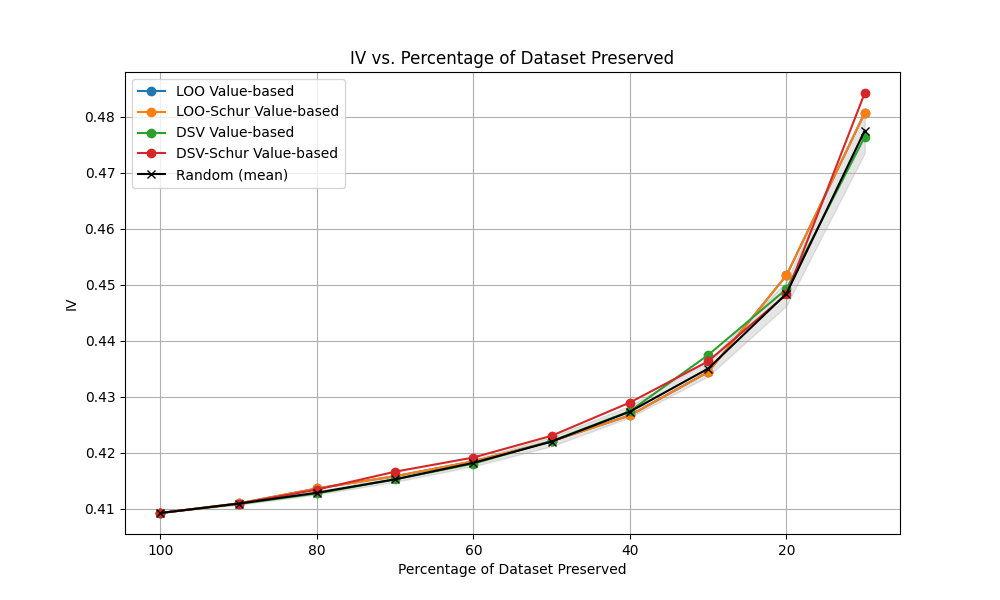}
    \caption{Data valuation impact on Integrated Variance with a kernel noise level of 0.01.}
    \label{fig:iv_low_noise}
\end{figure}

\subsection{Folktables}
\label{appendix:folktables}

The Folktables dataset, a variant of the Adult Income dataset, was used for predicting the numerical income values of individuals. For this dataset, the Truncated DSV algorithm was configured to perform only 10 iterations because of the dataset size (2000 data points).

Regarding the Folktables results, we noted that while LOO values were closely aligned, DSV exhibited disparities (Low Spearman coefficient) when the Schur complement was utilized. This indicates that the numerical instability, along with the choice of kernel and DSV parameters, can significantly influence empirical outcomes.

\begin{figure}[h!]
    \centering
    \includegraphics[width=0.8\textwidth]{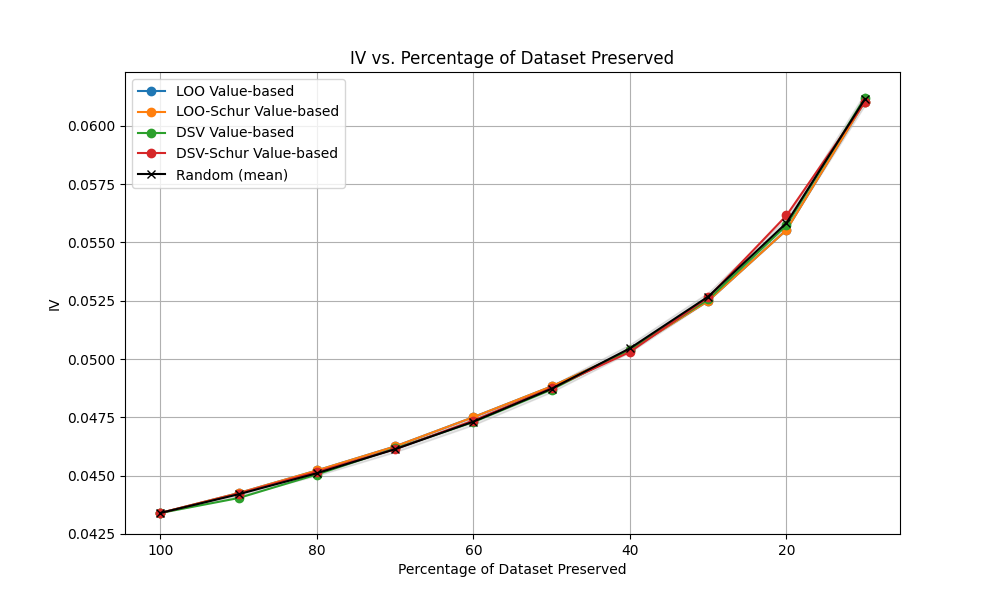}
    \caption{Data removal task results on the Folktables dataset, showing IV as a function of the dataset percentage preserved.}
    \label{fig:ft1_iv_vs_percentage}
\end{figure}

The data removal task on the Folktables did not conclusively show that removal based on value-based strategies had an advantage as shown in Figure~\ref{fig:ft1_iv_vs_percentage}, which suggests that kernel selection and DSV parameterization are critical to the success of these methods.

\end{document}